\pdfoutput=1

\documentclass[11pt]{article}

\usepackage{EMNLP2022}

\usepackage{times}
\usepackage{latexsym}

\usepackage[T1]{fontenc}

\usepackage[utf8]{inputenc}

\usepackage{microtype}

\usepackage{inconsolata}

\usepackage{todonotes}
\usepackage[portuguese, english]{babel}
\usepackage{color}
\usepackage{amsmath}
\usepackage{hyperref}
\usepackage{cleveref}
\usepackage{pgfplotstable}
\usepackage{tikz} 
\usepackage{graphicx} 
\usepackage{pgfplots}
\usepackage{booktabs}
\usepackage[acronym]{glossaries}
\usepackage{listings}
\usepackage{backnaur}

\usepackage{algorithm}
\usepackage{algpseudocode}
\usepackage{lipsum}
\usepackage{float}

\definecolor{backcolour}{rgb}{0.95,0.95,0.92}

\pgfplotsset{
    discard if not/.style 2 args={
        x filter/.code={
            \edef\tempa{\thisrow{#1}}
            \edef\tempb{#2}
            \ifx\tempa\tempb
            \else
                
            \fi
        }
    }
}

\usepgfplotslibrary{colorbrewer}
\pgfplotsset{
    cycle list/Paired-7,
    cycle multiindex* list={
        mark list*\nextlist
        Paired-7\nextlist
    },
}

\newacronym{sota}{SOTA}{state-of-the-art}
\newacronym{llm}{LLM}{large language model}
\newacronym{dl}{DL}{deep learning}
\newacronym{ir}{IR}{intermediate representation}
\newacronym{nl}{NL}{natural language}
\newacronym{em}{EM}{exact match}
\newacronym{ex}{EX}{execution match}
\newacronym{mda}{MDA}{model driven architecture}
\newacronym{ted}{TED}{tree edit distance}
\newacronym{ast}{AST}{abstract syntax tree}
\newacronym{cfg}{CFG}{context-free grammar}
\newacronym{seq2seq}{Seq2Seq}{sequence to sequence}
\newacronym{qm}{QM}{question match}
\newacronym{im}{IM}{interaction match}

\title{T5QL: Taming language models for SQL generation}

\author{Samuel Arcadinho, David Apar\'icio, 
\textbf{Hugo Veiga, António Alegria} \\
Outsystems \\
\{samuel.arcadinho, david.aparicio, hugo.veiga, antonio.alegria\}@outsystems.com}

\begin{document}
\maketitle
\begin{abstract}

Automatic SQL generation has been an active research area, aiming at streamlining the access to databases by writing natural language with the given intent instead of writing SQL. Current \gls*{sota} methods for semantic parsing depend on \glspl*{llm} to achieve high predictive accuracy on benchmark datasets. This reduces their applicability, since \glspl*{llm} requires expensive GPUs. Furthermore, \gls*{sota} methods are ungrounded and thus not guaranteed to always generate valid SQL. Here we propose T5QL, a new SQL generation method that improves the performance in benchmark datasets when using smaller LMs, namely \mbox{T5-Base}, by \mbox{$\approx13$pp} when compared against \gls*{sota} methods. Additionally, T5QL is guaranteed to always output valid SQL using a context-free grammar to constrain SQL generation. Finally, we show that dividing semantic parsing in two tasks, candidate SQLs generation and candidate re-ranking, is a promising research avenue that can reduce the need for large LMs.

\end{abstract}

\section{Introduction}

Automated code generation has long been considered one the fundamental tasks in computer science~\cite{pnueli1989synthesis}.
Recently, \gls*{dl} methods for code generation have been proposed which overcome the lack of flexibility of more traditional approaches~\cite{le2020deep}. Some \gls*{dl} approaches can act as code completion tools~\cite{svyatkovskiy2020intellicode, chen2021evaluating} while others can use \gls*{nl} as input to generate code~\cite{yin2017syntactic}, i.e., semantic parsing~\cite{kamath2018survey}. The latter methods are particularly helpful for developers that are not proficient in all programming languages that are part of their development pipeline. For example, a developer might be familiar with the controller language (e.g., Python) but unfamiliar with the database access language (e.g., SQL).

Generating SQL from \gls*{nl} is challenging because the \gls*{nl} query might be ambiguous (e.g., columns from different tables can have the same name). Futhermore, obtaining labelled pairs of NL queries to SQL is hard, time-consuming, and requires labellers that are proficient in SQL. In recent years, benchmark datasets have been used by developers to evaluate their methods, namely Spider~\cite{raffel2019exploring} and CoSQL~\cite{yu2019cosql}.

\begin{figure}[t]
  \begin{center}
    \begin{tikzpicture}[font=\small]
  \begin{axis}[
      width=\linewidth, 
      grid=major, 
      grid style={dashed,gray!30}, 
      legend style={nodes={transform shape}},
      ymin=0.4,
      ymax=0.80,
      xmin=0.8, xmax=8.2,
      xlabel=\textbf{beam size}, 
      ylabel=\textbf{exact match accuracy},
      every axis plot/.append style={thick, line width=0.5mm},
      legend pos=south west,
      legend columns=2,
      legend style={draw=none,fill=none, style={column sep=0.05cm, line width=0.1pt}},
      legend cell align={left},
      compat=1.11,
    ]
    
    \addplot 
    table [discard if not={type}{Ranker}, x=beam_size, y=em, col sep=comma]{data/em_beam_plot.csv}; 
    
    \addplot 
    table [discard if not={type}{T5-Base}, x=beam_size, y=em, col sep=comma]{data/em_beam_plot.csv}; 
    
    \addplot+[mark=none, samples=4, domain=1:8, dashed] {0.666};
    
    \addplot+[mark=none, samples=4, domain=1:8, dotted] {0.755};


    \addlegendentry{T5QL-Base}
    \addlegendentry{T5-Base}
    \addlegendentry{PICARD-Base}
    \addlegendentry{PICARD-3B}
    
  \end{axis}
\end{tikzpicture}
    \caption{Exact-match accuracy of the highest scoring prediction as a function of beam size on the Spider development set. Our method, T5QL, significantly improves upon T5-Base and is superior to PICARD-Base. PICARD-3B remains the \gls*{sota} for very large LMs, i.e., PICARD-3B uses T5-3B which is $\approx 13x$ larger than T5-Base.
    Results for PICARD-Base and PICARD-3B are straight (dashed) lines since \citet{scholak2021picard} only report results in the setting using database content for a single point, namely beam search with 4 beams.\label{fig:key_results}}
  \end{center}
\end{figure}
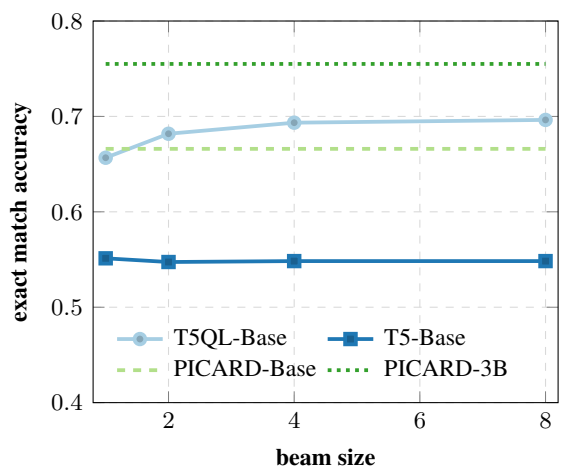
PICARD~\cite{scholak2021picard} is the current \gls*{sota} method, i.e., the highest ranked method on Spider. It is built on top of T5~\cite{raffel2019exploring}, a general purpose \gls*{llm}. As proven by \citet{merrill2021provable}, \glspl*{llm} are ungrounded and thus can generate any token at any given step, which may result in invalid SQL; thus, to improve upon T5, 
PICARD 
prunes the search tree in order to avoid generating invalid SQL. However, since PICARD fully prunes branches during beam search, it is not guaranteed to always generate an answer. 
Another major issue with PICARD 
is that 
it needs a very large LM to achieve good performance: PICARD gets $\approx$~75.5\% \gls*{em} accuracy in Spider's development set when using T5-3B, but only $\approx$~66.6\% when using the smaller T5-Base.

Here, we propose T5QL, a novel SQL generation method that achieves 69.3\% \gls*{em} on Spider development set using T5-Base instead of the $\approx$13x larger T5-3B.
T5QL uses constrained decoding to ensure that it always generates valid SQL, and it always generates an answer. Our main contributions are:

\begin{enumerate}
    \item Narrow the gap between large and small LMs (Figure~\ref{fig:key_results}). With beam size equal to 4 and using T5-Base, T5QL achieves 69.3\% \gls*{em} accuracy on Spider, versus 66.6\% obtained by PICARD. PICARD with T5-3B is still \gls*{sota} (75.5\%) but it requires much larger GPUs, which are expensive and thus not available for regular practitioners.
    \item Propose a constrained decoding method that always generates valid SQL, except for infrequent model hallucinations. In \Cref{sec:valid-sqls} we show one such case.
    \item Propose a novel ranker model for SQL generation. This model re-ranks the generator model's predictions after beam search, boosting EM on Spider for larger beam sizes (e.g., 8 beams) from 67.9\% to 69.6\%.
\end{enumerate}

The remainder of the paper is organized as follows.  Section~\ref{sec:sota} presents \gls*{sota} for SQL generation. Section~\ref{sec:method} describes T5QL's main components, namely constrained decoding and the ranker. Section~\ref{sec:experiments} shows our results. Finally, Section~\ref{sec:conclusion} concludes our work.

\section{State-of-the-art}\label{sec:sota}

Automated program generation has long been one of the major goals of computer science. 
Various program synthesis tools have been proposed that generate SQL from code fragments~\cite{cheung2012inferring} or pairs of input-output examples~\cite{orvalho2020squares}. However, code fragments might not be readily available if the developer does not write code or does not want to, and creating enough input-output examples for the program synthesis tool to be effective might be cumbersome. Other tools generate SQL from \gls*{nl} which is more developer-friendly~\cite{yaghmazadeh2017type}. 

The complexity of generating SQL from \gls*{nl} varies with the length and complexity of the SQL query and the size of the database schema. Thus, in order to properly evaluate and compare methods' performance, multiple benchmark datasets have been proposed, namely Spider~\cite{raffel2019exploring}, Spider-SSP~\cite{Shaw2021CompositionalGA}, and CoSQL~\cite{yu2019cosql}. We describe these benchmarks in detail in Section~\ref{sec:exp-datasets} and discuss how they relate to our research questions (enumerated at the start of Section~\ref{sec:experiments}).

The current \gls*{sota} for SQL generation (i.e., the methods that achieve the highest performance on benchmark datasets) comprises \gls*{dl} methods. \Gls*{dl} methods for code generation avoid the complexity of traditional program synthesis and, thus, are generally faster during generation~\cite{parisotto2016neuro, hayati2018retrieval, sun2019grammar}. 

RatSQL's authors argue that predicting SQL directly from NL is hard and can be made easier by instead predicting an \gls*{ir} that is more similar to NL than SQL is~\cite{wang2019rat, gan2021natural}. With this insight, they obtained \gls*{sota} results on Spider. However, their \gls*{ir} is not capable of representing all SQLs and, thus, for some queries the correct SQL is not obtainable, leading to a loss of \gls*{em} accuracy. 
Other approaches were built on top of RatSQL with good results~\cite{2021arXiv210109901Z, 2020arXiv201210309S, yu2020grappa}. SmBoP uses a semi regressive decoder instead of the autoregressive decoder employed by RatSQL, reducing the number of generation steps~\cite{rubin2020smbop}. 
RaSaP combines the decoder from SmBoP with the encoder strategy employed from RatSQL, but it uses a fine-tuned model as its encoder to improve its predictive capabilities~\cite{huang2021relation}.
One of the major disadvantages of these methods is that, since they use custom architectures, they cannot leverage pre-trained \glspl*{llm} in their decoding step. Being able to leverage \glspl*{llm} is beneficial since they can be used for multiple tasks. For example, \citet{Xie2022UnifiedSKGUA} unifies structured knowledge grounding tasks into a text-to-text format and are thus able to train the same model for different tasks. 

\begin{figure*}[t]
    \centering
    \includegraphics[width=0.99\linewidth]{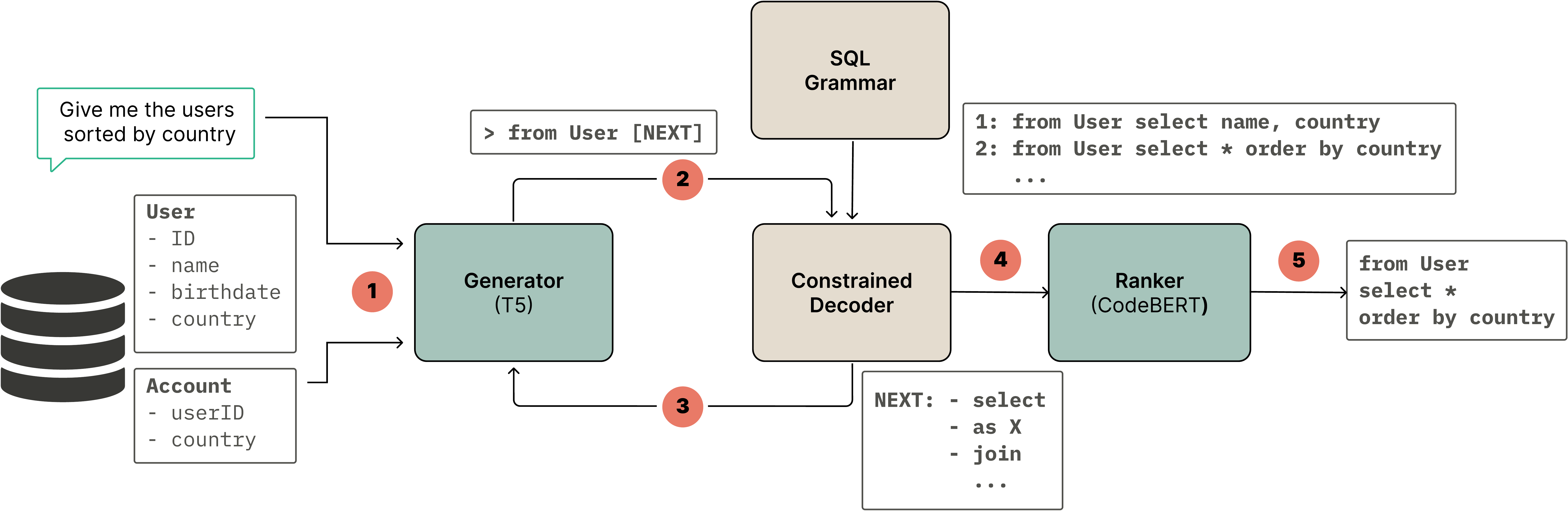}
    \caption{T5QL model architecture. T5QL receives as input an \gls*{nl} query and a database schema (step 1). Then, the generator model, T5, consults the constrained decoder to know which tokens are valid (step 2) and predicts the next token (step 3). This step is done iteratively. The generation is done using beam search, thus producing a set of $k$ candidates which are given as input to the ranker model (step 4). Finally, the ranker model ranks all candidates and a final prediction is outputted by T5QL (step 5). }
    \label{fig:model_architecture}
\end{figure*}

To the best of our knowledge, \citet{Shaw2021CompositionalGA} were the first to propose a method that uses an \gls*{llm}, namely T5, and evaluate it on Spider. They concluded that their method had good predictive capabilities, but sometimes generated syntactically incorrect SQL and had lower precision in out-of-distribution examples. Since T5 is ungrounded, it cannot be guaranteed to always generate valid SQL; the same is true for other \glspl*{llm}~\cite{merrill2021provable}. 
In order to address the issue, \citet{xiao-etal-2016-sequence} propose a method that constrains the output generation based on grammatical rules. They also compare a model trained with constraints and verify that using the constraints only during inference improves the model. More recently, \citet{de2020autoregressive} show that by constraining the output generation of a \gls{seq2seq} model, their method achieved \gls{sota} results when compared against more specialized systems, while reducing the memory footprint.

Targeting code generation specifically, \citet{scholak2021picard} propose PICARD, a method that constrains the model generation by removing wrong outputs during beam search. By doing so, PICARD is the current \gls*{sota} in the Spider benchmark. However, they report that PICARD did not generate any SQL for 2\% of the queries.   ~\citet{poesia2022synchromesh} improve \gls*{llm} performance in the few-shot setting by introducing two components, one that selects the examples to be given to the model and another that constrains the generation of syntactically correct SQL. However, fine-tuned models (e.g., PICARD) still perform better in the general task than their model, which was trained in the few-shot setting.

In this work, we use one model to generate candidates, a \emph{generator}, and another to re-rank them, a \emph{ranker}. This choice is motivated by recent work~\cite{chen2021evaluating} where the authors show that a re-ranking method boosted performance for code generation. Regarding semantic parsing more concretely, \citet{Ye2022RNGKBQAGA} use a ranker model to select candidates, and then a fine-tuned model generates the final output; their model shows good generalization capabilities and outperforms previous methods for question answering on knowledge graphs. More recently, \citet{rankgen} argue that when current \glspl*{llm} are given a prefix prompt they can often generate text that is incoherent with the prefix. They propose a ranker model that scores the generator's candidates for an input prefix and obtain results that outperform earlier models in both automatic and human evaluation.

\section{Method}~\label{sec:method}

We start this section by presenting an overview of our method and its architecture (Figure~\ref{fig:model_architecture}). Then, we focus on each of its main components, namely constrained decoding and the ranker. Finally, we discuss the scoring function and evaluation metrics.

\subsection{Overview}

Our method outputs the corresponding SQL query for a given \gls*{nl} query and a database schema. The database schema comprises a list of tables and their respective columns. Figure~\ref{fig:model_architecture} shows a simple \gls*{nl} query, "Give me the users sorted by country", and a toy database schema with only two tables, \texttt{User} and \texttt{Account}. The generator, T5, receives the \gls*{nl} and the database schema as input and, starting with an empty string, it iteratively predicts the next token. However, unlike regular T5, the next token prediction is limited by the constrained decoder to only consider tokens that form a valid SQL query up to that point. For example, if the current query is \texttt{"from User"}, the next valid tokens include \texttt{"select"}, \texttt{"as X"}, and \texttt{"join"}, but do not include \texttt{"from"} or \texttt{"User"}. We discuss why we invert the \emph{from} and the \emph{select} statements in Section~\ref{sec:constrained-decoding}. We use beam search to generate multiple candidate queries, which are given as input to the ranker model.

\subsection{Constrained decoding}\label{sec:constrained-decoding}

We use constrained decoding to limit which tokens are considered by the generator to make the next token prediction. In order to enforce valid tokens, we build a \gls*{cfg} of SQL statements. Our constrained decoding method, described in \Cref{alg:constrain}, is similar to the one proposed by~\citet{poesia2022synchromesh}: for each decoding step, given the current generation $P$, T5QL finds the maximum parsable prefix $P^*$, this means that all SQL tokens in the prefix $P^*$ have valid syntax (lines 2--5). Then, using the lookahead feature of the parser, T5QL tokenizes all possible suffixes for $P^*$ and adds them to trie $T$ (lines 6--10). Finally, T5QL computes possible generation tokens by searching the possibles suffixes for $P$ in $T$ (lines 11--12).

\begin{algorithm}[h]
  \caption{Constrained decoding}\label{alg:constrain}
  \begin{algorithmic}[1]
    \Procedure{NextToken}{$P$, $T$} \Comment{$P$ is the current SQL generation and $T$ the current trie}
      \State{$P^* \gets \Call{FindParsablePrefix}{P}$}
      
      \State{$S \gets \Call{GetParserState}{P^*}$}
      
      \State{$N \gets \Call{ParserNextTokens}{S}$}
      
      \State{$N^* \gets \Call{FilterWrongTokens}{S, N}$}
      
      \For {$ n $ in $ N^* $}
        \State{$C \gets P^* + n$}
        
        \State{$C^T \gets \Call{SentenceTokenizer}{C}$}
      
        \State{$T \gets \Call{AddToTrie}{T, C^T}$}
        
      \EndFor 
      
      \State{$P^T \gets \Call{SentenceTokenizer}{P}$}
      
      \State{$\Return{\Call{GetChildren}{T, P^T}}$}
      
    \EndProcedure
  \end{algorithmic}
\end{algorithm}

We note that, while our grammar is context free, our constrained decoding method uses context to make decisions: \textsc{FilterWrongTokens} (line 5 of \Cref{alg:constrain}) constrains the SQL generation by only allowing the generation of columns that are defined in the \emph{from} statement and by mapping table aliases to the original tables. We should point out that, while this is currently not performed by our method, we could extend constrained decoding to enforce more rules, such as only allowing tables to be joined using valid foreign keys or limiting the \emph{where} statement to only have conditions that have the proper return type given the column types (e.g., if a column "\texttt{X}" is of type \emph{string}, "\texttt{X > 10}" is not a valid generation).

Next, we focus on the grammar. For brevity, we only show higher-level statements below; the entire grammar is shown in our Codalab page\footnote{\url{https://worksheets.codalab.org/worksheets/0x0049b642db90440e9eaf9cf6a850b4c9}}. Statements inside square brackets indicate that they are optional (e.g., a SQL query can have an empty \emph{where} statement).

\begin{bnf*}
\bnfprod{sql}{\bnfpn{expr}}\\
\bnfprod{expr}{\bnfpn{query} \bnfor}\\
\bnfmore{\bnfpn{expr} \bnfsp \bnfts{union} \bnfsp \bnfpn{expr} \bnfor} \\
\bnfmore{\bnfpn{expr} \bnfsp \bnfts{intersect} \bnfsp \bnfpn{expr} \bnfor} \\
\bnfmore{\bnfpn{expr} \bnfsp \bnfts{except} \bnfsp \bnfpn{expr}} \\
\bnfprod{query}{\bnfts{from} \bnfsp \bnfpn{from-expr}} \\
\bnfmore{\bnfts{select} \bnfsp \bnfpn{select-expr}} \\
\bnfmore{[\bnfts{where} \bnfsp \bnfpn{where-expr}]} \\
\bnfmore{[\bnfts{group by} \bnfsp \bnfpn{groupby-expr}]} \\
\bnfmore{[\bnfts{having} \bnfsp \bnfpn{having-expr}]} \\
\bnfmore{[\bnfts{order by} \bnfsp \bnfpn{orderby-expr} ]} \\
\bnfmore{[\bnfts{limit} \bnfsp \bnfpn{limit-expr}]} \\
\end{bnf*}

Our grammar only supports SQL \emph{select} statements since our focus are queries that retrieve data from a database. These \emph{select} statements can be a single query or contain subqueries joined by \emph{unions}, \emph{intersects}, and \emph{excepts}. We note that the \emph{from} and the \emph{select} statements are inverted. This is done because, besides restricting T5 to only generate syntactically correct SQL, we also restrict it to only generate SQL with valid table names (i.e., tables that exist in the database schema) and valid column names (i.e., columns that exist in the database schema for the given table). To restrict the generation to only valid columns, it is helpful to first know the valid tables, which are obtained in the \emph{from} statement. Thus, T5QL first parses the \emph{from} statement and stores the selected tables; then, when the \emph{select} statement is parsed, T5QL already knows what columns are valid since they had to appear in the selected tables (e.g., from the example from Figure~\ref{fig:model_architecture}, if the current query is \texttt{"from User select"}, then \texttt{"user.ID"} and \texttt{"user.name"} are valid token predictions while \texttt{"account.country"} and \texttt{"account.userId"} are not).

For a given query and database schema pair, we augment the grammar shown previously with two extra rules specifying the valid tables and the valid columns. For the example from Figure~\ref{fig:model_architecture}, we would add the following production rules:

\begin{bnf*}
\bnfprod{table-name}{\bnfts{user} \bnfor \bnfts{account}} \\
\bnfprod{column-name}{\bnfts{user.id} \bnfor \bnfts{user.name} \bnfor} \\
\bnfmore{\bnfts{user.birthdate} \bnfor} \\
\bnfmore{\bnfts{user.name} \bnfor} \\
\bnfmore{\bnfts{user.country} \bnfor} \\
\bnfmore{\bnfts{account.userId} \bnfor} \\
\bnfmore{\bnfts{account.country}} \\
\end{bnf*}

When a table has an an alias, we add one expression for the alias and another for the original table table (e.g., for a column "\texttt{alias1.columnA}", we add two expressions to the ⟨column-name⟩ rule: "\texttt{alias1.columnA}" and "\texttt{tableX.columnA}", assuming that \texttt{alias1} corresponds to \texttt{tableX}).

The grammar is given as input to the Lark parser\footnote{\url{https://github.com/lark-parser/lark}}. We use Lark since it is one of the fastest parsers for Python, and it includes a look-ahead feature that we require. 

\subsection{Ranker}\label{sec:ranker}

We use beam search to generate a set of $k$ candidate queries and employ a ranker model to choose the best option among the $k$ candidates. We hypothesize that splitting the task of SQL generation into two tasks, (1) SQL candidates generation and (2) SQL candidate ranking, boosts the performance of the complete task since each model is only focused on a simpler task.

We use a trained generator model to generate the dataset to train the ranker model. The T5 model described in Section~\ref{sec:constrained-decoding} samples 16 SQL queries for each input (\gls*{nl} query and database schema pair) in the training dataset using beam search. From the 16 generated SQLs we sample the 12 with lowest \gls*{ted} (discussed in Section~\ref{sec:eval_metrics}) to guarantee that we select hard negative examples. If the generator model does not predict the correct SQL in any of the 12 SQLs samples, we discard the one with the highest \gls*{ted} and add the correct SQL as one of the samples. Using the same sampling strategy (i.e., based on \gls*{ted}), we sample an additional two SQLs from the training dataset pertaining to the same database as the input, for a total of 14 SQLs for each input. 

For the ranker model, we fine-tune CodeBERT \cite{2020arXiv200208155F} in a cross encoder setting. The ranker is given pairs of NL and SQL and predicts the probability of the pairs being correct, i.e., the SQL corresponding to the NL. We also append the terminals found in the \gls*{nl} using the method proposed by \citet{lin2020bridging} to the final \gls*{nl} (e.g., for the \gls*{nl} "People from 'France'", the \gls*{nl} is transformed into "People from 'France' | France").

\subsection{Scoring Function}

Similarly to \citet{yee-etal-2019-simple}, we compute the final prediction score for a given input by combining the generator's probability score with the ranker's probability score using the linear combination shown in Equation \ref{eq:score}, where $t$ is the length of the SQL, and $\lambda$ is a tunable weight. In order to compare the generator's probability $p(y|x)$ with the ranker's probability $p(x, y)$, we scale the generator's probability by $t$.

\begin{equation}
    \frac{1}{t} \log{p(y|x) + \lambda \log{p(x,y)}}
    \label{eq:score}
\end{equation}

\subsection{Evaluation metrics}\label{sec:eval_metrics}

The most commonly used evaluation metrics for SQL comparison are \gls*{em} and \gls*{ex}. \gls*{em} checks if two SQLs are syntactically equivalent, while \gls*{ex} checks if running two SQLs yields the same output. While desirable, \gls*{ex} is more computationally expensive than \gls*{em} since it requires running the SQL statements, which might not even be possible if we do not have access to the database content. When measuring the method's performance, it is also relevant to highlight if it also predicts terminal values or not; T5QL generates the full SQL query, including terminal values.

Since \gls*{em} is binary, its value might not be very informative for the user nor the model. Partial matches sub-divide the comparison to only portions of the SQL statement, such as the \emph{from} clause or the  \emph{where} clause. Thus, one SQL prediction might be wrong in multiple parts of the query, and this more granular information can be useful to improve the model. However, these measures are still coarse; thus, we use the \gls*{ted} in some experiments (namely in the ranker) when we want more information on the difference (or distance) between two SQLs. 

In order to compute the \gls*{ted} between two SQL statements, we transform each SQL statement into a tree and use APTED\footnote{\url{https://github.com/DatabaseGroup/apted}} to compute the TED between two trees. Due to SQL's semantics, we first normalize the SQL to a canonical representation (e.g., sort the list of tables in the \emph{select} alphabetically, transform \emph{left joins} into \emph{right joins}). Then modify APTED to guarantee that the TED is meaningful (e.g., the cost of removing a terminal and column name should be the same).

\section{Experiments}\label{sec:experiments}

We start by describing the experimental setup in \Cref{sec:exp-setup}. Then, we detail each dataset and the relevant evaluation metrics in \Cref{sec:exp-datasets}. Then, each subsequent section (\Cref{sec:q1}--\ref{sec:q3}) tries to answer each of the following research questions:


\begin{enumerate}\bfseries
    \item[Q1.] Does constrained decoding improve the generator's performance?
    \item[Q2.] Does T5QL have compositional generalization capabilities?
    \item[Q3.] Does T5QL generalize to the conversational setting?
    \item[Q4.] Instead of using a very large generator, can we improve performance using a ranker?
\end{enumerate}

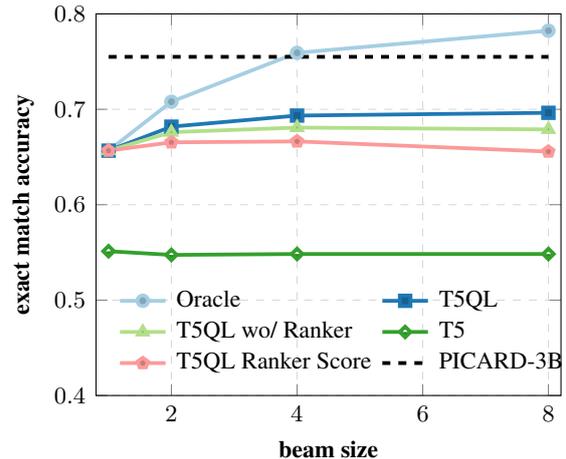
\begin{figure}[t]
  \begin{center}
    \begin{tikzpicture}[font=\small]
  \begin{axis}[
      width=\linewidth, 
      grid=major, 
      grid style={dashed,gray!30}, 
      legend style={nodes={transform shape}},
      ymin=0.4,
      ymax=0.80,
      xmin=0.8, xmax=8.2,
      xlabel=\textbf{beam size}, 
      ylabel=\textbf{exact match accuracy},
      every axis plot/.append style={thick, line width=0.5mm},
      legend pos=south west,
      legend columns=2,
      legend style={draw=none,fill=none, style={column sep=0.05cm, line width=0.1pt}},
      legend cell align={left},
      compat=1.11,
    ]

    \addplot 
    table [discard if not={type}{Oracle}, x=beam_size, y=em, col sep=comma]{data/em_beam_plot.csv};
    
    \addplot 
    table [discard if not={type}{Ranker}, x=beam_size, y=em, col sep=comma]{data/em_beam_plot.csv}; 
    
    \addplot 
    table [discard if not={type}{T5QL-Base}, x=beam_size, y=em, col sep=comma]{data/em_beam_plot.csv}; 
    
    \addplot 
    table [discard if not={type}{T5-Base}, x=beam_size, y=em, col sep=comma]{data/em_beam_plot.csv}; 
    
    \addplot 
    table [discard if not={type}{Ranker Only}, x=beam_size, y=em, col sep=comma]{data/em_beam_plot.csv}; 
    
    \addplot+[mark=none, samples=4, domain=1:8, dashed, black] {0.755};
    

    \addlegendentry{Oracle}
    \addlegendentry{T5QL}
    \addlegendentry{T5QL wo/ Ranker}
    \addlegendentry{T5}
    \addlegendentry{T5QL Ranker Score}
    \addlegendentry{PICARD-3B}
    
  \end{axis}
\end{tikzpicture}
    \caption{\gls*{em} accuracy in Spider's development set by beam size. All methods use T5-Base as their LM except for PICARD-3B which uses T5-3B. The performance of PICARD-3B is shown as a straight line since the authors only report results on the development set using database content for a single point (beam search with 4 beams). The Oracle plot shows the performance ceiling for T5QL, i.e., the performance of T5QL using a perfect ranker that always outputs the correct SQL if the generator offers it as one of the candidates after beam search.} \label{fig:componets_influence}
  \end{center}
\end{figure}

\subsection{Experimental setup}\label{sec:exp-setup}

For our experiments we use a G4DN Extra Large AWS machine, which has an NVIDIA T4 Tensor Core GPU installed and 4 CPU-cores. We make our code available in our public Codalab page\footnote{\url{https://worksheets.codalab.org/worksheets/0x0049b642db90440e9eaf9cf6a850b4c9}}\footnote{We will make the code available in github after the blind review process is finalized.}. 

\subsection{Datasets}\label{sec:exp-datasets}

We evaluate T5QL on three benchmark datasets: Spider~\cite{raffel2019exploring}, Spider-SSP~\cite{Shaw2021CompositionalGA}, and CoSQL~\cite{yu2019cosql}.

Spider comprises 10,181 \gls*{nl} and database schemas pairs, on 200 different database schemas. Evaluation on Spider consists of two main leaderboards: \gls*{ex} with terminal values and \gls*{em} without terminal values. At the time of writing, PICARD is the current \gls*{sota} method on both leaderboards. 

Spider-SSP is a different splitting of the Spider dataset, with the aim of testing compositional generation instead of cross-database generalization, i.e., in the original Spider data split, a database schema seen in train is not seen in eval or test. Splits in the Spider-SSP dataset are made in three different fashions: random split, a split based on source length, and a split based on Target
Maximum Compound Divergence (TMCD). The goal here is to evaluate if the model can have good performance on queries that it has not seen in training.

While Spider consists of a single NL and domain model pair mapped into a single SQL query, CoSQL consists of a conversational dataset with multiple iterations of NL plus data model being mapped to a SQL query. The goal of CoSQL is to simulate a user progressively exploring a data model. CoSQL contains 4,298 interactions and $\approx 12,000$ questions, on the same 200 data models used in Spider. Evaluation is done using \gls*{em} without terminal values and reported using two different metrics: \gls*{qm} and \gls*{im}. \gls*{qm} evaluates if all SQLs are correctly predicted, while \gls*{im} evaluates if the questions for the same interaction are correctly predicted.

\subsection{Q1. Constrained decoding}\label{sec:q1}

LMs are unconstrained and thus can generate any token at any given time. For SQL generation, LMs may generate SQL that are syntactically incorrect, which impact their performance. 

Here, we compare the performance of an unconstrained LM against T5QL without the ranker component. Both methods use the same LM, namely T5-Base, and are trained using the same training configuration; the only difference is that T5QL uses constrained decoding as described in \Cref{sec:constrained-decoding}. Both methods serialize the
database schema as a string and append it to the
source sequence similarly to \citet{suhr2020exploring}. Similarly to \citet{scholak2021picard}, we train both methods for a maximum of 512 training epochs with mini batch size of 5, 205 gradient accumulation steps, with a learning rate of $1e^{-4}$, and an adafactor optimizer with epsilon set as $1e^{-6}$. We evaluate the models using beam search with 1, 2, 4, and 8 beams. Contrary to \citet{scholak2021picard}, we report results for a batch size of $1025$ instead of $2048$ since it lead to better results in our case.

\Cref{fig:componets_influence} shows the performance of several methods and those results are discussed in this subsection and in the next ones. All methods use T5-Base as its \gls*{llm}, except for PICARD which is the current \gls*{sota} and uses T5-3B, a much larger LM. 

From \Cref{fig:componets_influence}, we observe that T5 achieves $\approx55.1\%$ \gls*{em} accuracy using one beam, and its performance does not improve with the beam size. Our method, T5QL, without the ranker component (i.e., T5QL wo/ Ranker in \Cref{fig:componets_influence}) achieves $\approx65.7\%$ \gls*{em} accuracy using one beam, a gain of $\approx10.6$pp, which is a relative gain of $\approx19.2\%$. Using 2 and 4 beams, we improve T5QL's performance to $\approx67.6\%$ and $\approx68\%$, a gain of $\approx1.9$pp and $\approx2.3$pp, respectively, when compared against T5QL using only one beam. We observe a loss of performance when using 8 beams. These results highlight the advantage of using constrained decoding for SQL generation: by using a \gls*{cfg} to guarantee that the LM always generates valid SQL, we improve the model's performance.

\subsection{Q2. Compositional generalization}

Compositional generalization of \glspl*{llm} has attracted attention in recent years. \citet{Shaw2021CompositionalGA} propose Spider-SSP, a dataset that can be used to measure the compositional generalization of SQL generation methods. In this section we use  Spider-SSP to evaluate if constraint decoding increases the compositional generalization capabilities of T5QL.

\citet{Shaw2021CompositionalGA} already reported that T5-Base model struggles in most splitting strategies, particularly when using length-based split and TMCD split; we reproduce those results in \Cref{tab:spider_ssp} in rows T5-Base and T5-3B. We note that, in their experiments, the predicted SQL follows the convention of predicting first the \emph{select} statement and then the \emph{from} statement. As discussed in \Cref{sec:constrained-decoding}, T5QL first predicts the \emph{from} statement and then the \emph{select} statement. Thus, we evaluate two different models: T5-Base, which is similar to the model evaluated by \citet{Shaw2021CompositionalGA}, and T5QL-Base wo/ CD which is T5QL without the constrained decoding component (and without the ranker). We compare these models against T5QL-Base and T5-3B; the latter also predicts the \emph{select} statement first.

We observe that T5QL-Base wo/ CD obtains significantly higher \gls*{em} than T5-Base, namely for the TMCD split where there is a gain of 22pp, which is a 52\% relative gain. These results highlight that predicting the tables before predicting the columns seems to help the model. This result corroborates the results obtained by \citet{lin2020bridging}, which use a representation similar to ours. We also verify that T5QL-Base slightly, but consistently, improves upon the results obtained by T5QL-Base wo/ CD for all splitting strategies, namely in TMCD where there is a gain of 2pp. Finally, we conclude that our strategy narrows the performance gap between the performance of methods using small LMs (i.e., T5-Base) and very large LMs (i.e., T5-3B) by comparing the performance of T5QL-Base against T5-3B.

\begin{table}[b]
\small
\begin{filecontents*}{EMNLP 2022/data/spider_ssp.csv}
model,tmcd,template,length,random
T5-Base,42.3,45.3,42.5,76.5
T5QL-Base wo/ CD,64.4424131627057,58.3029197080292,50.6398537477148,84.7349177330896
T5QL-Base,65.9049360146252,61.1313868613139,54.3875685557587,85.7404021937843
T5-3B,69.6,64.8,56.7,85.6
\end{filecontents*}
\pgfplotstabletypeset[
    precision=1,
	col sep=comma,
	columns={model, random, template, length, tmcd},
	every head row/.style={%
        before row={\toprule
            & \multicolumn{4}{c}{\textbf{Spider-SSP}} \\ 
            \cmidrule{2-5}
        },
        after row=\midrule
    },
    every last row/.style={after row=\bottomrule},
	columns/model/.style={column name=Model, column type={l},string type},
	columns/tmcd/.style={column name=TMCD},
	columns/template/.style={column name=Templ.},
	columns/length/.style={column name=Len.},
	columns/random/.style={column name=Rand.},
]{EMNLP 2022/data/spider_ssp.csv}

\caption{\gls*{em} accuracy in the Spider-SSP dataset using different splitting strategies. T5QL-Base wo/ CD (i.e., without constrained decoding) and T5QL-Base adopt the strategy of predicting SQL with the \emph{from} statement before the \emph{select} statement, while T5-Base and T5-3B do the opposite, which is the default. }
\label{tab:spider_ssp}
\end{table}

\subsection{Q3. Generalize to conversational setting}\label{sec:q4}

Often users might want to explore their data without having to write SQL. Thus, a conversational setting where user's iteratively ask questions to an AI is particularly interesting. \citet{yu2019cosql} propose a dataset comprised of multiple question-SQL pairs, each consisting of several user interactions. They evaluate SQL generation methods using \gls*{qm} and \gls*{im}. In this section we use CoSQL to evaluate if constrained decoding increases the performance of T5SQL in the conversational setting.

We observe gains of $\approx 7.9\%$ and $\approx 5.5\%$ in \gls*{qm} and \gls*{im}, respectively, when we add constrained decoding to T5QL-Base. We observe that PICARD-3B is still \gls*{sota} for the task, but the gap is significantly narrower. This is further evidence that constrained decoding can improve the performance of LMs in multiple SQL generation tasks.

\begin{table}[t]
\centering
\small
\begin{filecontents*}{EMNLP 2022/data/cosql_result.csv}
Model,QM,IM
T5QL-Base wo/ CD ,42.8,14.8
T5QL-Base,50.7,20.3
PICARD - 3B,56.9,24.2
\end{filecontents*}
\pgfplotstabletypeset[
    precision=1,
	col sep=comma,
	columns={Model, QM, IM},
	every head row/.style={%
        before row={\toprule
            & \multicolumn{2}{c}{\textbf{CoSQL}} \\ 
            \cmidrule{2-3}
        },
        after row=\midrule
    },
    every last row/.style={after row=\bottomrule},
	columns/Model/.style={column name=Model, column type={l},string type},
	columns/Question Match/.style={column name=QM},
	columns/Interaction Match/.style={column name=IM},
]{EMNLP 2022/data/cosql_result.csv}
\caption{\gls*{qm} and \gls*{im} in the CoSQL development set.}
\label{tab:cosql}
\end{table}

\subsection{Q4. Ranker}\label{sec:q3}
Current \gls*{sota} methods, such as PICARD, use beam search to find the best candidate and output it as the final prediction. Here we test whether we can boost predictive performance by, instead of using the beam-selected best candidate candidate as the final prediction, having a ranker that finds the best candidate among the list of candidate predictions found by the generator.

Our first step to validate this hypothesis is to run beam search for multiple beam sizes $k$, namely 1, 2, 4, and 8, and measure the accuracy@k. In our setting, the accuracy@k can be regarded as an \emph{oracle} ranker than can always find the correct candidate if it is present in the list of candidate generations. From \Cref{fig:componets_influence} we observe that this \emph{oracle} could achieve $78.2\%$ \gls*{em} accuracy with 8 beams, surpassing the performance of PICARD-3B but using T5-Base instead of T5-3B, which is highly desirable due to T5-3B's expensive nature in terms of GPU costs. Thus, our goal here is to build a ranker model that can boost the performance of T5QL without the ranker (T5QL wo/ Ranker in \Cref{fig:componets_influence}) and approximate it to the \emph{oracle}'s performance.

We note that the ranker model should be of a comparable size to the generator, i.e., fit in the same GPU. Otherwise, the advantage of using a small LM as the generator is lost since we assume that the practitioner has hardware that can fit the larger ranker, which might not be true. Here we use T5-Base as the generator and CodeBERT as the ranker, which are of comparable size.

To train the ranker model, we first create a dataset following the steps described in \Cref{sec:eval_metrics}. Then, we fine-tune a CodeBERT model for 50,000 training steps, using a batch size of 32 and 1 gradient accumulation step, with a $1e^{-3}$ learning rate and an AdamW optimizer with weight decay of $1e^{-2}$ and a linear schedule with warmup of $1\%$ of the steps. We use Equation~\ref{eq:score} to score the generated SQL; we conduct hyperparameter tuning for $\lambda$ and conclude that $\lambda = 2e^{-2}$ performs best. 

We analyze if combining the generator's score with the ranker's score is superior to using each of the score's individually. From \Cref{fig:componets_influence} we conclude that combining the ranker model's score with the generator model's score (i.e., the T5QL plot) improves the best \gls*{em} from $67.9\%$ to $69.3\%$ when compared against T5QL without the ranker score. Furthermore, we also observe that using only the ranker score (i.e., the T5QL Ranker Score plot) leads to a drop in performance even when compared against T5QL wo/ Ranker. This effect is more noticeable for larger beam sizes, which indicates that the ranker model struggles to differentiate the correct SQL from the wrong SQL.

From these experiments, we conclude that the ranker boosts the performance of the generator. However, the ranker's score needs to be combined with the generator's score to guarantee that the ranker's score does not completely dominate the generator's predictions. We should also note that there is a very large gap between our ranker and the oracle, which leaves room for future research to improve the ranker model. We believe that this a promising line of research that can further narrow the gap between the performance between small LMs and large LMs.

Finally, we run T5SQL on Spider's test set and obtain 66.8\% \gls*{ex} and 65.9\% \gls*{em}. These results rank among the top-10 best models in terms of \gls*{ex}, and as the 22nd best in terms of \gls*{em}\footnote{\url{https://yale-lily.github.io/spider}}, whilst using small models. Small models have the advantage of being less computationally expensive and allowing more easily for the use of ensemble methods.

\section{Conclusion and discussion}\label{sec:conclusion}

Here we put forward T5QL, a new method for SQL generation with \gls*{sota} performance on benchmark datasets when using small LMs. T5QL uses constrained decoding to improve predictive performance and also to guarantee that the generated SQL is always valid. Futhermore, we complement the generator model with a ranker model that is capable of choosing the best candidate SQL from a pool of a few candidates.

In this work we do not address topics that are relevant for future studies, namely the limitations of current benchmarks (e.g., the database schemas in Spider are small when compared to data models of real-world applications) and evaluation metrics (i.e., how to measure the impact to the user of giving them an invalid SQL? e.g., if the model predicts "MyUsername" instead of "MyUserName" and the user does not notice, we might have added a bug to their code). Furthermore, we believe that building more capable ranker models is a promising line of research that can greatly boost SQL generation methods' performance, removing the need to use very large LMs.



\bibliography{references}
\bibliographystyle{acl_natbib}
\newpage
\appendix
\section{SQL generation analysis}\label{appendix:indepth}

In this section we analyse in detail the predictions generated by T5QL. In \Cref{sec:valid-sqls} we measure how often T5QL outputs valid SQLs and give an example of one invalid SQL. In \Cref{sec:existing-table-names} we show an example of how constraining column name generation can boost performance.

\subsection{Valid SQL generation}\label{sec:valid-sqls}

First, we check if T5QL using constrained decoding can still generate unparsable SQL. We obtain T5QL-Base's predictions in Spider's development set for beam sizes of 1, 2, 4 and 8. We observe that:
\begin{itemize}
    \item T5QL never generates an unparsable SQL for the top-1 beam when the beam size $> 1$.
    \item Invalid SQL is generated when the LM (i.e., T5) enters a loop, as can be seen in Listing~\ref{lst:unparsable_sql}. Since the SQL length is limited, T5QL outputs the incomplete (and invalid) SQL. The loop, even if abnormal, is valid SQL syntax, e.g, an average of averages.
    \item For larger beam sizes (e.g, 8) we saw that the aforementioned model hallucinations are mainly present in the lower scored beams.
\end{itemize}

\begin{lstlisting}[
    frame=tb,
    backgroundcolor=\color{backcolour},
    basicstyle={\small\ttfamily},
    numbers=none,
    breaklines=true,
    breakatwhitespace=false,
    showstringspaces=false,
    caption={Invalid SQL generated by T5QL. For space concerns we abbreviate the generated SQL.},label={lst:unparsable_sql}
]
from stadium select name, capacity order by avg( avg( avg( avg( avg( avg( avg( avg( avg( avg( avg( avg( avg( avg( avg( avg( avg( avg( avg( avg( max( avg( min( min( min( min( min( max( max( max( max( max( max( max( max( max( max( max( max( max( max( max( max( max( max( max( max( max( max( max( ...
\end{lstlisting}

Next, we analyse whether model size reduces the number of invalid SQL generated by T5QL. We obtain the predictions in Spider's development set using T5QL-Base and T5QL-Large with and without constrained decoding. We report results of the four methods using 4 beams.

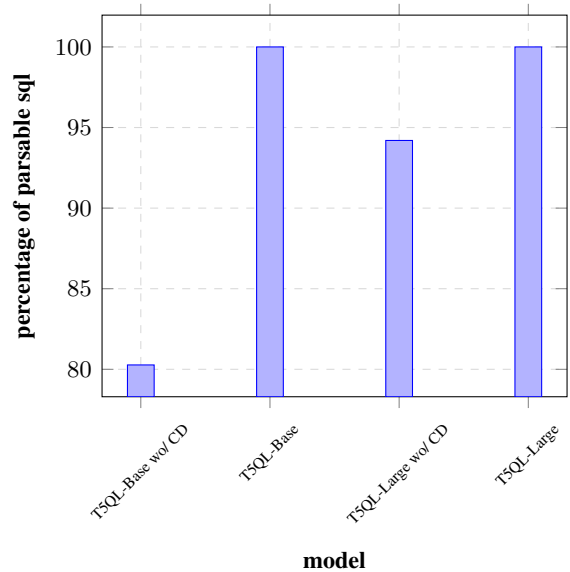
\begin{figure}[t]
  \begin{center}
    \pgfplotstableread[col sep=comma]{data/analysis_parsed_sql.csv}\datatable

\begin{tikzpicture}[font=\small]
  \begin{axis}[
      ybar,
      width=\linewidth, 
      grid=major, 
      grid style={dashed,gray!30}, 
      xlabel=\textbf{model}, 
      ylabel=\textbf{percentage of parsable sql},
      xtick=data,
      xticklabels from table={\datatable}{model},
      xticklabel style={font=\tiny, rotate=45},
      compat=1.11,
    ]
    
    \addplot table [x expr=\coordindex, y=Percentage] {\datatable};

    
    
  \end{axis}
\end{tikzpicture}
    \caption{Percentage of parsable SQL, in the Spider's development set, in each model configuration. All methods use beam search with 4 beams and we report results for the first beam.} \label{fig:correct_sqls}
  \end{center}
\end{figure}

We observe that increasing the size of the model also increases the ability of the model to generate parsable SQL: T5QL-Base wo/ CD generates $\approx 20\%$ invalid SQLs, while T5QL-Large wo/ CD generates only $\approx5\%$ invalid SQLs (\Cref{fig:correct_sqls}). Notice, however, that 5\% is still a substantial amount of invalid SQLs. On the other hand, when using constrained decoding, T5QL always produces valid SQLs when considering only the top-1 beam of beam search with 4 beams; this is true for both T5QL-Base and T5QL-Large.

\subsection{Enforce existing table and column names}\label{sec:existing-table-names}

Finally, we analyse what is the impact of constraining the table and column names during SQL generation. When T5QL does not constrain column and table names, it can generate examples such as the one in Listing~\ref{lst:invalid-sql-ex} where "\texttt{song\_id}" is a made-up column name that does not exist in the schema. When constraining column and table names, T5QL always generates existing column and table names, and, in this case, predicts the correct SQL (Listing~\ref{lst:valid-sql-ex}).

\begin{lstlisting}[
    frame=tb,
    backgroundcolor=\color{backcolour},
    basicstyle={\small\ttfamily},
    numbers=none,
    breaklines=true,
    breakatwhitespace=false,
    showstringspaces=false,
    caption={Invalid SQL generated by T5QL wo/ CD. In this case the T5QL generated an non-existing column.}
    \label{lst:invalid-sql-ex}
]
    from singer as t1 join singer_in_concert as t2 on t1.song_id = t2.song_id select t1.name, count( * ) group by t1.song_id
\end{lstlisting}

\begin{lstlisting}[
    frame=tb,
    backgroundcolor=\color{backcolour},
    basicstyle={\small\ttfamily},
    numbers=none,
    breaklines=true,
    breakatwhitespace=false,
    showstringspaces=false,
    caption={Valid and correct SQL generated by T5QL with CD for the same example as Listing~\ref{lst:invalid-sql-ex}.}
    \label{lst:valid-sql-ex}
]
    from singer as t1 join singer_in_concert as t2 on t1.singer_id = t2.singer_id select t1.name, count( * ) group by t1.singer_id
\end{lstlisting}

\section{Larger models}

\begin{figure}[t]
  \begin{center}
    \begin{tikzpicture}[font=\small]
  \begin{axis}[
      width=\linewidth, 
      grid=major, 
      grid style={dashed,gray!30}, 
      legend style={nodes={transform shape}},
      ymin=0.4,
      xlabel=\textbf{beam size}, 
      ylabel=\textbf{exact match accuracy},
      every axis plot/.append style={thick, line width=0.5mm},
      legend pos=south west,
      legend columns=2,
      legend style={draw=none,fill=none, style={column sep=0.05cm, line width=0.1pt}},
      legend cell align={left},
      compat=1.11,
    ]

    \addplot  table [discard if not={model}{T5QL-Base wo/ Constraints}, x=k, y=em, col sep=comma]{data/analysis_base_large.csv};
    
    \addplot  table [discard if not={model}{T5QL-Large wo/ Constraints}, x=k, y=em, col sep=comma]{data/analysis_base_large.csv};
    
    \addplot  table [discard if not={model}{T5QL-Base}, x=k, y=em, col sep=comma]{data/analysis_base_large.csv};
    
    \addplot  table [discard if not={model}{T5QL-Large}, x=k, y=em, col sep=comma]{data/analysis_base_large.csv};
    
    \addplot  table [discard if not={model}{T5QL-Base + Ranker}, x=k, y=em, col sep=comma]{data/analysis_base_large.csv};
    
    \addplot table [discard if not={model}{T5QL-Large + Ranker}, x=k, y=em, col sep=comma]{data/analysis_base_large.csv};


    \addlegendentry{Base wo/ CD}
    \addlegendentry{Large wo/ CD}
    \addlegendentry{Base wo/ R}
    \addlegendentry{Large wo/ R}
    \addlegendentry{Base}

    \addlegendentry{Large}
   
    
  \end{axis}
\end{tikzpicture}
    \caption{Comparison of \gls*{em} accuracy in Spider's development between different model configurations. The "Base" model refers to T5QL-Base with constraint decoding and reranking; models "wo/ CD" are the models without constraint decoding nor reranking, whereas models "wo/ R" are the models without reranking.} \label{fig:compare_base_large}
  \end{center}
\end{figure}

Experiments shown previously for our proposed method, T5QL, always used T5-Base as the base LM. We make this choice since our focus is to show that small LMs can have good performance even when compared against very large LMs. Nevertheless, evaluating if the proposed techniques, namely constrained decoding and reranking, scale to larger LMs is an interesting research questions. Thus, here we evaluate whether constrained decoding and reranking improve the performance of T5SQL-Large.

From Figure~\ref{fig:compare_base_large} we observe that the performance of T5QL-Base (i.e., Base) is superior to T5-Large (i.e., Large wo/CD) for 2--4 beams. When we add the constrained decoding component to T5-Large (i.e., Large wo/ R), the performance is significantly superior. This results highlights the importance of adding constrained decoding for SQL generation. However, we do not observe gains of adding the reranker model to T5-Large (i.e., Large), which we observed in T5-Base. This might indicate, as we pointed out in Section~\ref{sec:ranker}, that finding better reranking strategies is an interesting research path.

We do not include results for T5QL-3B since our main goal in this work is to increase performance using multiple smaller components (e.g., a small SQL generation model and a reranker) and domain-aware techniques (e.g., constrained decoding) instead of relying on very large models. Furthermore, computing results for T5-3B is very costly in terms of money and time.

\end{document}